%% file: cvpr.tex
\pdfoutput=1
\documentclass[final]{cvpr}
\usepackage[percent]{overpic}
\usepackage{times}
\usepackage{epsfig}
\usepackage{graphicx}
\usepackage{amsmath}
\usepackage{amssymb}
\usepackage{subcaption}
\usepackage{booktabs} % for professional tables
\newcommand{\R}{\mathbb{R}}
 \usepackage{booktabs}
\usepackage{multirow}
\newif\ifdraft
\drafttrue
 \usepackage[table,xcdraw]{xcolor}

\usepackage[pagebackref=true,breaklinks=true,colorlinks,bookmarks=false]{hyperref}

\def\cvprPaperID{9903} % *** Enter the CVPR Paper ID here
\def\confYear{CVPR 2021}
\begin{document}

%%%%%%%%% TITLE
\title{Revamping Cross-Modal Recipe Retrieval with Hierarchical Transformers and Self-supervised Learning}

\author{Amaia Salvador \:\:\:  Erhan Gundogdu \:\:\:  Loris Bazzani \:\:\:  Michael Donoser
\and
\small{Amazon} \\
\small{\{asalvada, eggundog, bazzanil, donoserm\}@amazon.com}}

\maketitle

%%%%%%%%% ABSTRACT

\input{sections/0_abstract.tex}
\input{sections/1_intro.tex}
\input{sections/2_related_work.tex}
\input{sections/3_method.tex}

\input{sections/4_experiments.tex}

\input{sections/5_conclusions.tex}

{\small
\bibliographystyle{ieee_fullname}
\bibliography{egbib}
}

\end{document}

%% file: sections/0_abstract.tex
\begin{abstract}
Cross-modal recipe retrieval has recently gained substantial attention due to the importance of food in people's lives, as well as the availability of vast amounts of digital cooking recipes and food images to train machine learning models. In this work, we revisit existing approaches for cross-modal recipe retrieval and propose a simplified end-to-end model based on well established and high performing encoders for text and images. We introduce a hierarchical recipe Transformer which attentively encodes individual recipe components (titles, ingredients and instructions). Further, we propose a self-supervised loss function computed on top of pairs of individual recipe components, which is able to leverage semantic relationships within recipes, and enables training using both image-recipe and recipe-only samples. We conduct a thorough analysis and ablation studies to validate our design choices. As a result, our proposed method achieves state-of-the-art performance in the cross-modal recipe retrieval task on the Recipe1M dataset. We make code and models publicly available\footnote{\scriptsize{\url{https://github.com/amzn/image-to-recipe-transformers}}}. %Recent methods in this domain made significant advances by leveraging existing trends in learning joint embeddings, generative adversarial methods and attention mechanisms. %outperforming previous LSTM-based methods commonly used for this task 
%reaching a medR of 3.0 for 10k-sized rankings
\end{abstract}

%% file: sections/1_intro.tex
\section{Introduction}
\label{intro}
% Intro on food
Food is one of the most fundamental and important elements for humans, given its connection to health, culture, personal experience, and sense of community.
With the development of the Internet and the rise of social networks, we witnessed a substantial surge in digital recipes that are shared online by users. Designing powerful tools to navigate such large amounts of data can support individuals in their cooking activities to enhance their experience with food, and has thus become an attractive research field \cite{min2019survey}. Often times, digital recipes come along with companion content such as photos, videos, nutritional information, user reviews, and comments. The availability of such rich large scale food datasets has opened the doors for new applications in the context of food computing  \cite{salvador2017learning, ofli2017saki, li2020picture}, one of the most prevalent ones being cross-modal recipe retrieval, where the goal is to design systems that are capable of finding relevant cooking recipes given a user submitted food image. Approaching this challenge requires developing models in the intersection of natural language processing and computer vision, as well as being able to deal with unstructured, noisy, and incomplete data.

\begin{figure}[t]
    \centering
    \includegraphics[width=\columnwidth]{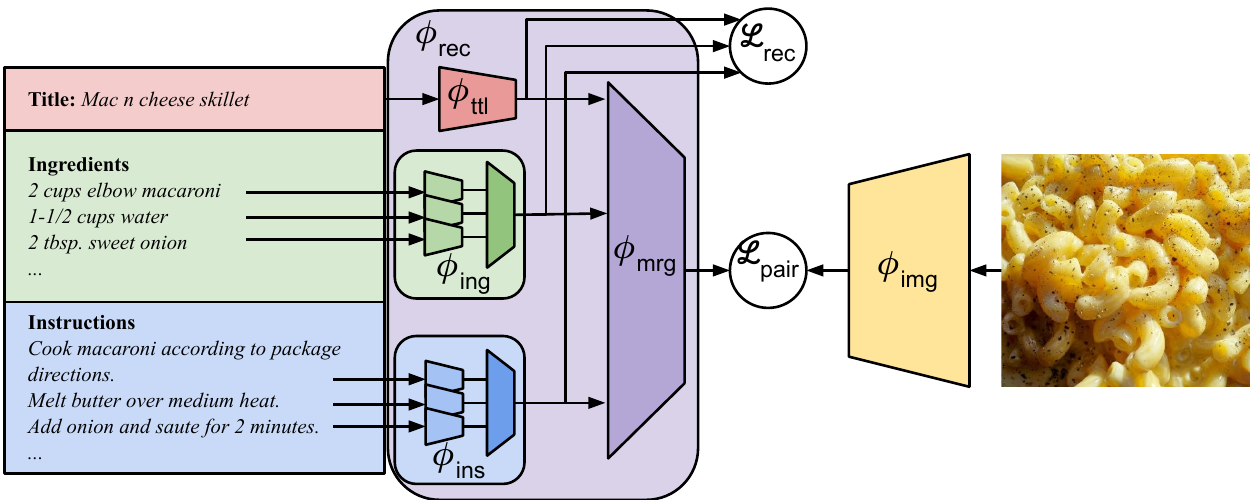}
    \caption{\textbf{Model overview.} Our method is composed of three distinct parts: the image encoder $\phi_{img}$, the recipe encoder $\phi_{rec}$, and the training objectives $\mathcal{L}_{pair}$ and $\mathcal{L}_{rec}$. }
    \label{fig:introfig}
    \vspace{-2mm}
\end{figure} 

% Learning joint representations
In this work, we focus on learning joint representations for textual and visual modalities in the context of food images and cooking recipes.
Recent works on the task of cross-modal recipe retrieval~\cite{salvador2017learning, carvalho2018cross, chen2018deep, wang2019learning, zhu2019r2gan} have introduced approaches for learning embeddings for recipes and images, which are projected into a joint embedding space that is optimised using contrastive or triplet loss functions. Advances were made by proposing complex models and loss functions, such as cross-modal attention~\cite{fu2020mcen}, adversarial networks~\cite{zhu2019r2gan, wang2019learning}, the use of auxiliary semantic losses~\cite{salvador2017learning, carvalho2018cross, wang2019learning, zhu2019r2gan,  wang2019learning}, multi-stage training~\cite{salvador2017learning, fain2019dividing}, and reconstruction losses~\cite{fu2020mcen}.
These works are either complementary or orthogonal to each other while bringing certain disadvantages such as glueing independent models~\cite{salvador2017learning, fain2019dividing}, which needs extra care, relying on a pre-trained text representations~\cite{salvador2017learning, carvalho2018cross, chen2018deep, wang2019learning, zhu2019r2gan, fain2019dividing} and complex training pipelines involving adversarial losses~\cite{zhu2019r2gan, wang2019learning}. 
% concept 1: simplicity of the model and compare to literature & comprehensive analysis 
In contrast to previous works, we revisit ideas in the context of cross-modal recipe retrieval and propose a \emph{simplified} end-to-end joint embedding learning framework that is plain, effective, and straightforward to train. Figure~\ref{fig:introfig} shows an overview of our proposed approach. 

Unlike previous works using LSTMs to encode recipe text~\cite{wang2015recipe,chen2016deep, chen2017cross, salvador2017learning, carvalho2018cross}, we introduce a recipe encoder based on Transformers~\cite{vaswani2017attention}, with the goal of obtaining strong representation for recipe inputs (i.e. titles, ingredients, and instructions) in a bottom-up fashion (see Figure.~\ref{fig:introfig}, left). Following recent works in the context of text summarization \cite{zhang-etal-2019-hibert, liu2019hierarchical}, we leverage the structured nature of cooking recipes with hierarchical Transformers, which encode lists of ingredients and instructions by extracting sentence-level embeddings as intermediate representations, while learning relationships within each textual modality. Our experiments show superior performance of Transformer-based recipe encoders with respect to their LSTM-based counterparts that are commonly used for cross-modal recipe retrieval.%The idea is to leverage on the power of self-attention in the transformer network to find relationships within each textual modality, \emph{e.g.}, the correlation between different instructions steps (or ingredients).

% For the image modality, we take advantage of well-established convolutional networks.

% issue 2: paired data
Training joint embedding models requires cross-modal paired data, i.e., each image must be associated to its corresponding text. In the context of cross-modal recipe retrieval, this involves quadruplet samples of pictures, title, ingredients, and instructions. Such a strong requirement is often not fulfilled when dealing with large scale datasets curated from the Web, such as Recipe1M \cite{salvador2017learning}. Due to the unstructured nature of recipes that are available online, Recipe1M largely consists of text-only samples, which are often either ignored or only used for pretraining text embeddings. In this work, we propose a new self-supervised triplet loss computed between embeddings of different recipe components, which is optimised jointly and end-to-end with the main triplet loss computed on paired image-recipe embeddings (see $\mathcal{L}_{rec}$ in Figure~\ref{fig:introfig}). The addition of this new loss allows us to use both paired and text-only data during training, which in turn improves retrieval results. Further, thanks to this loss, embeddings from different recipe components are aligned with one another, which allows us to recover (or \textit{hallucinate}) them when they are missing at test time.

% concept 2: deal with unpaired data 
%Our model is able to deal with the absence of images associated to the recipe as showed in Fig.~\ref{fig:introfig}.
%We propose a loss function that is composed by the following triplet losses.
%1) The standard loss that is applied on paired data~\cite{wang2016learning}. 
%2) A self-supervised loss that considers recipes without images, that is defined by looking at pairs of textual modalities (e.g., ingredient-title).

%\TODO{expand point 2. }
%\AS{All works using pre-trained text representations from \cite{salvador2017learning} are implicitly using all the dataset. We are the first to do so in an end-to-end fashion by proposing a loss that is optimized together with the main loss for paired data.}

Our method encodes recipes and images using simple yet powerful model components, and is optimised using both paired and unpaired data thanks to the new self-supervised recipe loss. Our approach achieves state-of-the-art results on Recipe1M, one of the most prevalent dataset in the community. We perform an ablation study to quantify the contribution of each of the design choices, which range from how we represent recipes, the impact of our proposed self-supervised loss, and a state-of-the-art comparison. % while not sacrificing from computation or memory requirements.

%\TODO{add improvement numbers}
%\TODO{Amaia did a crazy exp analysis, we should put more emphasis on that here}

The \emph{contributions} of this work are the following. \textbf{(1)} We propose a recipe encoder based on hierarchical Transformers that significantly outperforms its LSTM-based counterparts on the cross-modal recipe retrieval task.
\textbf{(2)} We introduce a self-supervised loss term that allows our model to learn from text-only samples by exploiting relationships between recipe components.
% 3) We introduce self-supervised loss functions so that our model can learn also in the case when paired data is not available. 
\textbf{(3)} We perform extensive experimentation and ablation studies to validate our design choices (i.e. recipe encoders, image encoders, impact of each recipe component). 
\textbf{(4)} As a product of our analysis, we propose a simple, yet effective model for cross-modal recipe retrieval which achieves state-of-the-art performance on Recipe1M, with a medR of $3.0$, and Recall@1 of $33.5$, improving the performance of the best performing model  \cite{fain2019dividing} by 1.0 and 3.5 points, respectively.

%% file: sections/2_related_work.tex
\section{Related Work}
\label{relatedwork}

%\begin{figure*}[h!]
%    \centering
%    \includegraphics[width=\textwidth]{figures/embed.png}
%    \caption{Image-Recipe Joint Embedding. The image encoder transforms the image into a fixed sized representation $e_I$. Each of the components of the recipe is separately encoded with an LSTM-based encoder: the title is encoded by a single LSTM, while ingredients and instructions are encoded with a Hierarchical LSTM (see Figure \ref{fig:hlstm}). The outputs of the three encoders are averaged to produce the recipe embedding. The model is optimized using the standard triplet loss with margin using positive and negative pairs of image/recipe embeddings. }
%    \label{fig:jointembed}
%\end{figure*}

\subsection{Visual Food Understanding}

The computer vision community has made significant progress on food recognition since the introduction of new datasets, such as  Food-101 \cite{bossard2014food} and ISIA Food-500 \cite{min2020isia}).
Most works focus on food image classification \cite{Liu2016DDL,ofli2017saki,Ngo2017DLF,nu9070657,chen2017chinesefoodnet,Lee_2018_CVPR}, where the task is to the determine the category of the food image. %, for which progress has been analogous to the architectural advances in image recognition~\cite{he2016deep, xie2017aggregated}. 
Other works study different tasks such as estimating ingredient quantities of a food dish \cite{Chen2012quantities, li2020picture}, predicting calories \cite{im2calories}, or predicting ingredients in a multi-label classification fashion  \cite{chen2016deep, chen2017cross}. 
Since the release of multi-modal datasets such as Recipe1M \cite{salvador2017learning}, new tasks in the context of leveraging images and textual recipes have emerged. 
Several works proposed solutions that use image-recipe paired data for cross-modal recipe retrieval \cite{wang2019learning, chen2018deep, salvador2017learning, carvalho2018cross, wang2020cross}, recipe generation  \cite{salvador2019inverse, wang2020structure, chandu2019storyboarding, amac2019procedural,nishimura2019procedural}, image generation from a recipe \cite{zhu2020cookgan, pan2020chefgan} and question answering \cite{yagcioglu2018recipeqa}. 
%Understanding cooking using multi-modal cues has also been explored in the video domain. With the release of datasets such as Epic Kitchens \cite{Damen2018EPICKITCHENS}, and YouCook2\footnote{\url{http://youcook2.eecs.umich.edu/}}, HowTo100m \cite{miech2019howto100m}, recent works have proposed tasks such as modality alignment \cite{malmaud2015s}  or text to video retrieval \cite{miech20endtoend}. 
Our paper tackles the task of cross-modal recipe retrieval between food images and recipe text. 
In the next section, we focus on the specific contributions of previous works addressing this task, highlighting their differences with respect to our proposed solution.

%recipe generation from flowgraphs \cite{Hammond86,MoriMYS14,MoriMSYHFY14}, ingredient checklists \cite{Kiddon16}.

\subsection{Cross-Modal Recipe Retrieval}

Learning cross-modal embeddings for images and text is currently an active research area \cite{Karpathy:2017, gu2018look, huang2019acmm}. Methods designed for this task usually involve encoding images and text using pre-trained convolutional image recognition models and LSTM \cite{lstm} or Transformer \cite{vaswani2017attention} text encoders. %optimizing contrastive or triplet losses between embeddings of different modalities. 

% pretrained text embeddings vs learned ones
In contrast to short descriptions from captioning datasets  \cite{chen2015microsoft, young2014image, sharma2018conceptual}, cooking recipes are long and structured textual documents which are non-trivial to encode. 
Due to the structured nature of recipes, previous works proposed to encode each recipe component independently, using late fusion strategies to merge them into a fixed-length recipe embedding. 
Most works \cite{salvador2017learning, carvalho2018cross, wang2019learning, zhu2019r2gan, fain2019dividing} do so by first pre-training text representations (e.g. word2vec \cite{mikolov2013efficient} for words, skip-thoughts \cite{kiros2015skip} for sentences), training the joint embedding using these representations as fixed inputs. In contrast to these works, our approach resembles the works of \cite{chen2018deep, fu2020mcen} in that we use the raw recipe text directly as input, training the representations end-to-end.

% text encoder architectures, consensus in which recipe data and how to use it
In the literature of cross-modal recipe retrieval, there is still no consensus with regards to how to best utilise the recipe information, and which encoders to use to obtain representations for each component. First, most early works \cite{salvador2017learning, carvalho2018cross, wang2019learning, zhu2019r2gan,  wang2019learning} treat recipe ingredients as single words, which requires an additional pre-processing step to extract ingredient names from raw text (e.g. extracting \textit{salt} from \textit{1 tbsp. of salt}). Only a few works \cite{chen2018deep, fu2020mcen} have removed the need for this pre-processing step by using raw ingredient text as input. Second, it is worth noting that most works ignore the recipe title when encoding the recipe, and only use it to optimise auxiliary losses \cite{salvador2017learning, carvalho2018cross, wang2019learning, zhu2019r2gan,  wang2019learning}. Third, when it comes to architectural choices, LSTM encoders are the choice of most previous works in the literature, using single LSTMs to encode sentences (e.g. titles, categorical ingredient lists), and hierarchical LSTMs to encode sequences of sentences (e.g. raw ingredient lists or cooking instructions). In contrast with the aforementioned works, we propose to standardise the process of encoding recipes by (1) using the recipe in its complete form (i.e. titles, ingredients, and instructions are all inputs to our model), and (2) removing the need of pre-processing and pre-training stages by using text in its raw form as input. Further, we propose to use Transformer-based text encoders, which we empirically demonstrate to outperform LSTMs.

% more losses, complex architectures
While triplet losses are often used to train such cross-modal models, most works proposed auxiliary losses that are optimised together with the main training objective. Examples of common auxiliary losses include cross-entropy or contrastive losses using pseudo-categories extracted from titles as ground truth \cite{salvador2017learning, carvalho2018cross, wang2019learning, zhu2019r2gan,wang2019learning, fu2020mcen} and adversarial losses on top of reconstructed inputs \cite{zhu2019r2gan, wang2019learning, fu2020mcen}, which come with an increase of complexity during training. 
Other works have also proposed architectural changes such as incorporating self- and cross-modal-attention mechanisms \cite{fu2020mcen}. 
In our work, we err on the side of simplicity by using encoder architectures that are ubiquitous in the literature (namely, vanilla image encoders such as ResNets and Transformers), optimised with triplet losses.

%unpaired data
Finally, it is worth noting that while Recipe1M is a multi-modal dataset, only 33\% of the recipes contain images. Previous works \cite{salvador2017learning, carvalho2018cross, chen2018deep, zhu2019r2gan, wang2019learning} only make use of the paired samples to optimise the joint image-recipe space, while ignoring the text-only samples entirely \cite{fu2020mcen, wang2020cross}, or only using them to pre-train text embeddings \cite{salvador2017learning, carvalho2018cross, chen2018deep, wang2019learning, zhu2019r2gan, fain2019dividing}. In contrast, we introduce a novel self-supervised loss that is computed on top of the recipe representations of our model, which allows us to train with additional recipes that are not paired to any image in an end-to-end fashion.% without adding training or model complexity. 

%% file: sections/3_method.tex
\section{Learning Image-Recipe Embeddings}
\label{model}

%Recipes come in the textual form including a title, a list of ingredients, and a list of instructions along with images depicting the food dish. 
%We optimize a joint neural embedding space with an objective that encourages the embeddings of positive image-recipe pairs to be similar, while penalising high similarities between embeddings belonging to different samples.
%In the absence of pairs, we present a self-learning objective that is defined by looking at pairs of recipe components (e.g., ingredient-title).

We train a joint neural embedding on data samples from a dataset of size $N$: $\{(x_I^n,x_R^n)\}_{n=1}^{N} $. 
Each $n^{th}$ sample is composed of an RGB image $x_I$ depicting a food dish, and its corresponding recipe $x_R = (r_{ttl}, r_{ing}, r_{ins})$, composed of a title $r_{ttl}$, a list of ingredients $r_{ing}$, and a list of instructions $r_{ins}$. In case that the recipe sample is not paired to any image, $x_I^j$ is not available for the $j^{th}$ sample and only $x_R^j$ is used during training. Figure \ref{fig:introfig} shows an overview of our method. Images and recipes are encoded with $\phi_{img}$ and $\phi_{rec}$, respectively, and embedded into to the same space through $\mathcal{L}_{pair}$. We incorporate a self-supervised recipe loss $\mathcal{L}_{rec}$ acting on pairs of individual recipe components. We describe each of these components below. 
% \EG{Self-supervised loss can be very briefly mentioned here.}

\subsection{Image Encoder  $\phi_{img}$}
\label{ssec:image_encoders}
The purpose of the image encoder is to learn a mapping function $e_I^n = \phi_{img}(x_I^n)$ which projects the input image $x_I^n$ into the joint image-recipe embedding space. We use ResNet-50 \cite{he2016deep} initialised with pre-trained ImageNet \cite{krizhevsky2012imagenet} weights as the image encoder. We take the output of the last layer before the classifier and project it to the joint embedding space with a single linear layer to obtain an output of dimension $D = 1024$. We also experiment with ResNeXt \cite{xie2017aggregated} based models, as well as the recently introduced Vision Transformer (ViT) encoder \cite{dosovitskiy2020image}\footnote{We use the ViT-B/16 pretrained model from \url{https://rwightman.github.io/pytorch-image-models/}.}. 

\subsection{Recipe Encoder  $\phi_{rec}$}
\label{ssec:recipe_encoders}
The objective of the recipe encoder is to learn a mapping function $e_R^n = \phi_{rec}(x_R^n)$ which projects the input recipe $x_R^n$ into the joint embedding space to be  directly compared with the image $e_I^n$.
Previous works in the literature have encoded recipes using LSTM-based encoders for recipe components, which are either pre-trained on self-supervised tasks \cite{salvador2017learning, carvalho2018cross, wang2019learning, zhu2019r2gan}, or optimised end-to-end \cite{fu2020mcen,wang2020cross} with an objective function computed for paired image-recipe data.
Similarly, our model uses a specialised encoder for each of the recipe components (namely, title, ingredients, and instructions). We use three separate encoders to process sentences from the title, ingredients and instructions. In contrast to previous works, we propose to use Transformer-based encoders for recipes as opposed to LSTMs, given their ubiquitous usage and superior performance in natural language processing tasks.

\begin{figure}
    \centering
    \begin{subfigure}[t]{0.45\columnwidth}
        \includegraphics[width=\columnwidth]{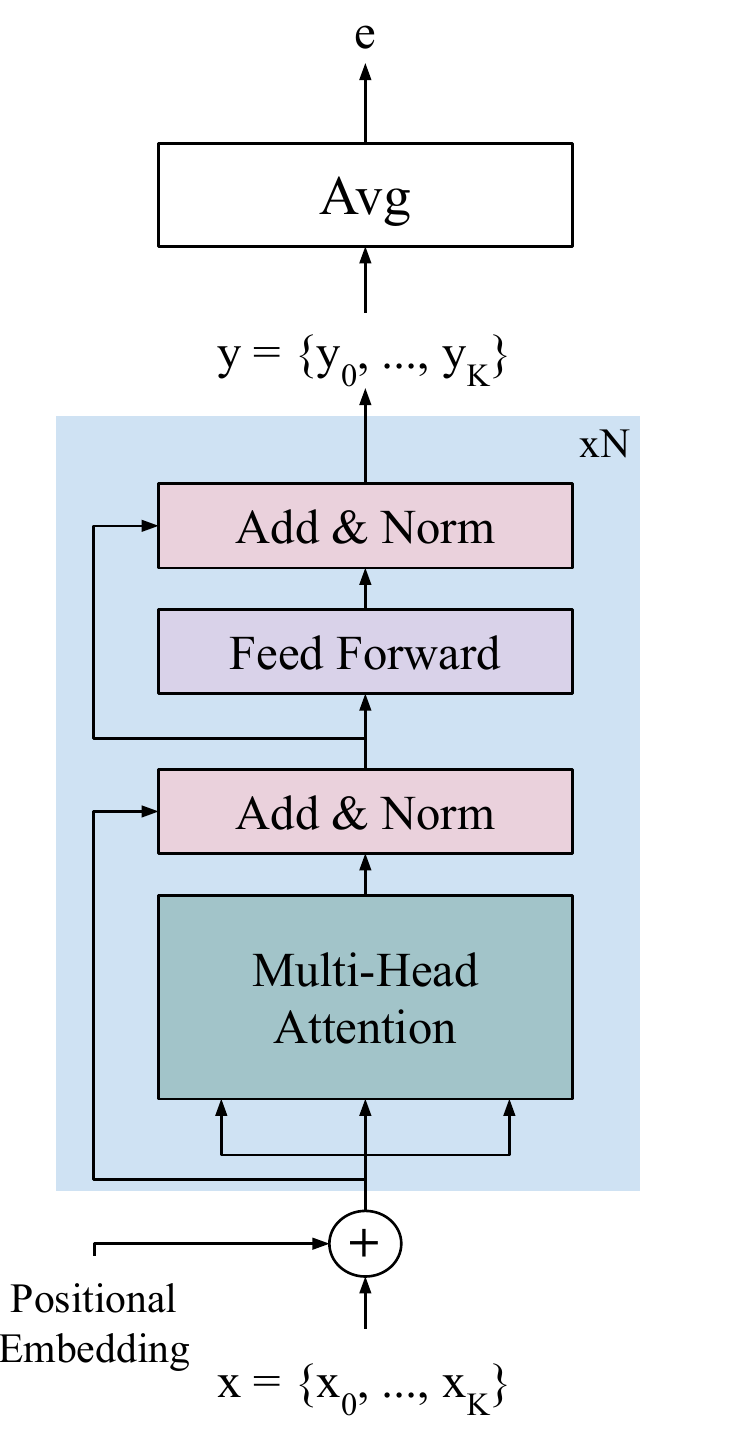}
        \caption{ $\mathrm{TR}$}
        \label{fig:tf_model}
    \end{subfigure}
    \begin{subfigure}[t]{0.45\columnwidth}
        \includegraphics[width=\columnwidth]{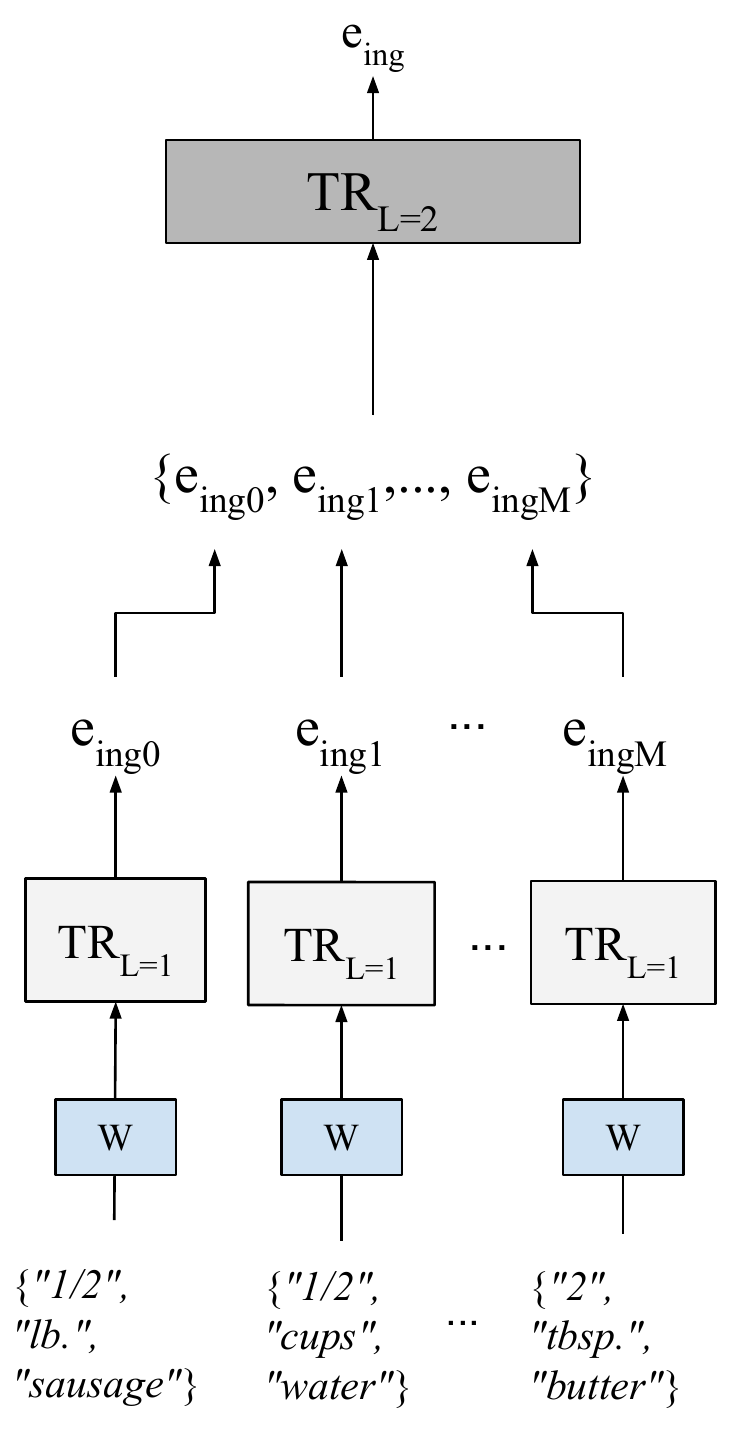}
        \caption{$\mathrm{HTR}$}
        \label{fig:h-tf_model}
    \end{subfigure}
    \caption{\textbf{(a) Transformer Encoder, TR:} Given a recipe sentence, our model encodes it into a fixed length representation using the Transformer encoder. \textbf{(b) Hierarchical Transformer Encoder, HTR:} For sequences of sentences (i.e. ingredients, or instructions), we use a hierarchical model, where a first Transformer encodes each sentence separately into a fixed sized vector, and a second Transformer encodes them into a single representation.}\label{fig:transformer_models}
\end{figure}

\textbf{Sentence Representation.} 
Given a sequence of word tokens $s = (w^0, ..., w^K)$, we seek to obtain a fixed length representation that encodes it in a meaningful way for the task of cross-modal recipe retrieval. 
The title consists of a single sentence, i.e. $r_{ttl} = s_{ttl}$, while instructions and ingredients are list of sentences, i.e. $r_{ing} = (s_{ing}^0, ..., s_{ing}^M)$ and $r_{ins} = (s_{ins}^0, ..., s_{ins}^O)$.
We take advantage of the training flexibility of Transformers~\cite{vaswani2017attention} for encoding sentences in the recipe. We encode each sentence with Transformer network of 2 layers of dimension $D = 512$, each with 4 attention heads, using a learned positional embeddings in the first layer. The representation for the sentence is the average of the outputs of the Transformer encoder at the last layer. Figure \ref{fig:tf_model} shows a schematic of our Transformer-based sentence encoder, $\mathrm{TR}(\cdot, \theta)$, where $\theta$ are the model parameters, which are different for each recipe component. Thus, we extract title embeddings as: $e_{ttl}=\phi_{ttl}(r_{ttl})=\mathrm{TR}(r_{ttl}, \theta_{ttl})$.

\textbf{Hierarchical Representation.} 
Both ingredients and instructions are provided as lists of multiple sentences. To account for these differences and exploit the input structure, we propose a hierarchical Transformer encoder, named $\mathrm{HTR}(\cdot, \theta)$, which we will use to encode inputs composed of sequences of sentences (see Figure ~\ref{fig:h-tf_model}). Given a list of sentences of length $M$, a first Transformer model $TR_{L=1}$ is used to obtain $M$ fixed-sized embeddings, one for every sentence in the list. Then, we add a second Transformer $TR_{L=2}$ with the same architecture (2 layers, 4 heads, $D=512$) but different parameters, which receives the list of sentence-level embeddings as input, and outputs a single embedding for the list of sentences. We use this architecture to encode both ingredients and instructions separately, using different sets of learnable parameters: $e_{ing}=\phi_{ing}(r_{ing})=\mathrm{HTR}(r_{ing}, \theta_{ing})$, and $e_{ins}=\phi_{ins}(r_{ins})=\mathrm{HTR}(r_{ins}, \theta_{ins})$.

The recipe embedding $e_R$ is computed with a final projection layer on top of concatenated features from the different recipe components: $e_R = \phi_{mrg}([e_{ing}; e_{ins}; e_{ttl}])$, where $\phi_{mrg}$ is a single learnable linear layer of $D=1024$, and  $[ \cdot ; \cdot ;]$ denotes embedding concatenation\footnote{We also experimented with embedding average instead of concatenation, which gave slightly worse retrieval performance.}. 

In order to compare with previous works~\cite{salvador2017learning, carvalho2018cross, wang2019learning, zhu2019r2gan,  wang2019learning}, we also experimented with LSTM \cite{lstm} versions of our proposed recipe encoder with the same output dimensionality $D = 512$, keeping the last hidden state as the representation for the sequence. 
%We thus experimentally show that our proposed strategy for representing recipe components using hierarchical Transformers makes a significant contribution to the retrieval performance compared to LSTM variants.

\subsection{Supervised Loss for Paired Data, $\mathcal{L}_{pair}$}
Inspired by the success of triplet hinge-loss objective for recipe retrieval \cite{carvalho2018cross, chen2018deep, wang2019learning, zhu2019r2gan}, we define the main component of our loss function as follows:
\begin{equation}
\mathcal{L}_{cos}(a, p, n) = \mathrm{max}(0, c(a, n) - c(a, p) + m)
\end{equation}
where $a$, $p$, and $n$ refer to the anchor, positive, and negative samples, $c(\cdot)$ is the cosine similarity metric, and $m$ is the margin (empirically set to $0.3$ for all triplet losses used in this work). 
In practice, we use the popular bi-directional triplet loss function~\cite{wang2019learning} on feature sets $a$ and $b$:
\begin{equation}
\begin{aligned}
\mathcal{L}_{bi}'(i, j) = {} 
& \mathcal{L}_{cos}(a^{n=i}, b^{n=i}, b^{n=j}) \\
& +   \mathcal{L}_{cos}(b^{n=i}, a^{n=i}, a^{n=j})
\end{aligned}
\end{equation}
where $a^{n=i}$ and $b^{n=i}$ are positive to each other, and $b^{n=j}$ and $a^{n=j}$  are negative to $a^{n=i}$ and $b^{n=i}$, respectively. We use the notation $n=i$ to denote same-sample embeddings (e.g. recipe and image embeddings from the same sample in the dataset) and $n = j$ for embeddings from a different sample $j$. During training, for a batch of size $B$, the loss for every sample $i$ is the average of all losses considering all other samples in the batch as negatives.
\begin{equation}
\mathcal{L}_{bi}(a^{n=i}, b^{n=i}, b^{n \neq i}, a^{n \neq i}) =\dfrac{1}{B} \sum_{j=0}^{B} \mathcal{L}_{bi}'(i,j) \delta(i,j), \label{eq:lossnegatives}
\end{equation}
%\vspace{-2mm}
where $\delta(i,j) = 1$ if $ i \neq j$ and $0$ otherwise. In the case that we have paired image-recipe data, we define the following loss by setting $a$ and $b$ to correspond to the image and recipe embeddings, respectively:
\begin{equation}
\mathcal{L}_{pair} = \mathcal{L}_{bi}(e_{I}^{n = i}, e_{R}^{n = i}, e_{R}^{n \neq i}, e_{I}^{n\neq i}) \label{eq:losspair}
\end{equation}
where $e_I = \phi_{img}(I)$, $e_R = \phi_{rec}(R)$ with ${e_I, e_R} \in \R^D$ are fixed sized representations extracted using the image and recipe encoders described in the previous sections.

%\begin{figure}[h]
%    \centering
%    \includegraphics[width=\columnwidth]{figures/recipe_loss.pdf}
%    \caption{Illustration of our proposed recipe loss $L_{recipe}$. Dashed arrows are identity projections.}
%    \label{fig:hlstm}
%\end{figure}
%\subsection{Beyond paired training data}
\subsection{Self-supervised Recipe Loss, $\mathcal{L}_{rec}$}
\label{ssec:additional_data}
%\EG{Title can be changed to something related to self-supervised loss if the ACME setting with the additional losses introduces an improvement. This is because we can sell it not with the use of additional data but with a novel use of self-supervision.}
%\amaia{This needs to be rephrased - brain dump at the moment}

%\EG{I would suggest that there should be two underlying motivations. (1) our loss with separate projections for each triplet loss is helpful even with the paired data because it makes indirect similarity relations and (2) allows to make use of unpaired data.}

% \textbf{Exploiting recipe-only samples}.

In the presence of unpaired data or partially available information, it is not possible to optimise Eq.~\ref{eq:losspair} directly. This is a rather common situation for noisy datasets collected from the Internet. In the case of Recipe1M, 66\% of the dataset consists of samples that do not contain images, i.e. only include a textual recipe.  In practice, this means that $e_I$ is missing for those samples, which is why most works in the literature simply ignore text-only samples to train the joint embeddings. However, many of these works~\cite{salvador2017learning, carvalho2018cross, chen2018deep, zhu2019r2gan,  wang2019learning} use recipe-only data to pre-train text representations, which are then used to encode text. While these works implicitly make use of all training samples, they do so in a suboptimal way, since such pre-trained representations are unchanged when training the joint cross-modal embeddings for retrieval. 

\begin{figure}
    \centering
        \includegraphics[width=0.8\columnwidth]{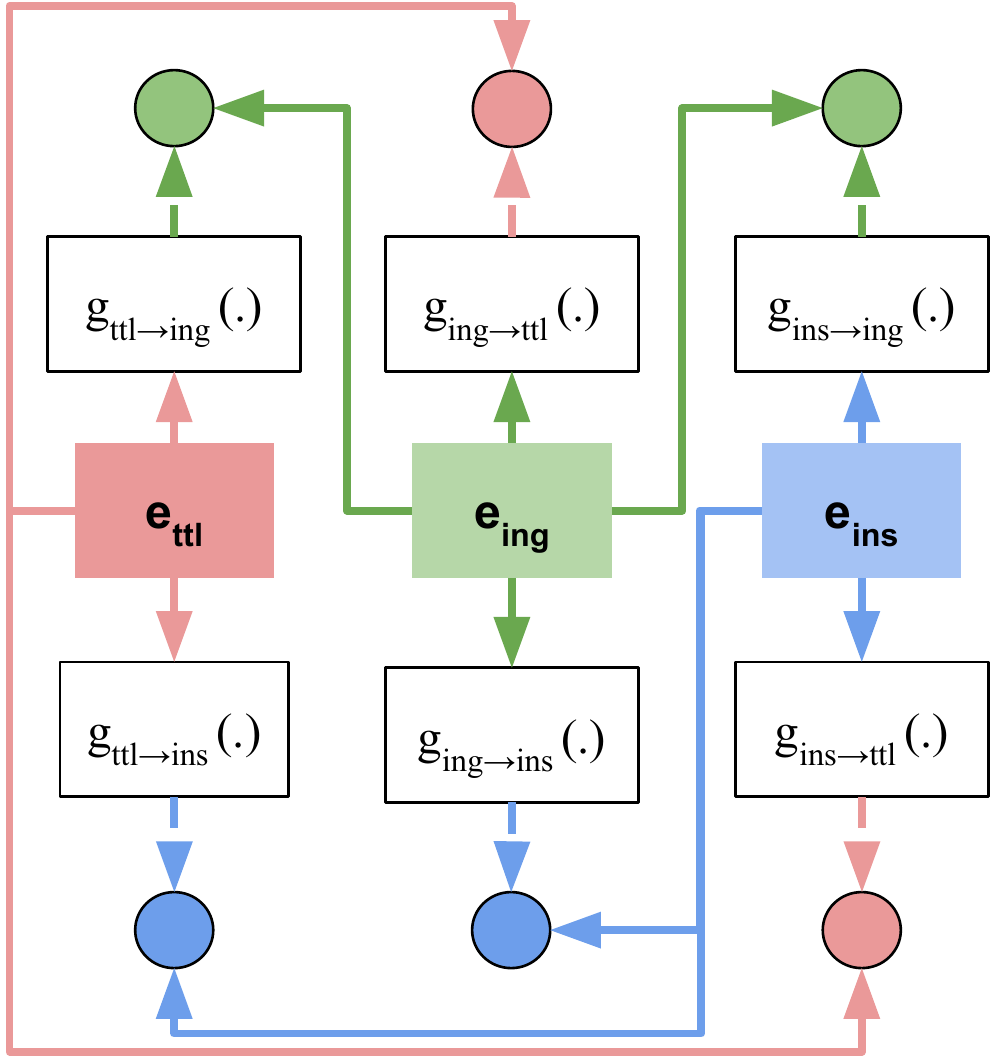}
    \caption{\textbf{Self-supervised recipe losses.}  Coloured dots denote loss terms computed for each recipe component. Each component embedding, e.g. $e_{ttl}$, is optimised to be close to the projected embeddings of the other two recipe components, namely: $g_{ing \rightarrow ttl}(e_{ing})$, and $g_{ins \rightarrow ttl}(e_{ins})$.}\label{fig:recipe_loss}
\end{figure}

%The motivation behind this loss is to leverage text-only samples to improve the recipe representation by means of enforcing embeddings from individual recipe components to be close in the intermediate space. 
In this paper, we propose a simple yet powerful strategy to relax the requirement of relying on paired data when training representations end-to-end for the task of cross-modal retrieval. Intuitively, while the individual components of a particular recipe (i.e. its title, ingredients, and instructions) provide complementary information, they still share strong semantic cues that can be used to obtain more robust and semantically consistent recipe embeddings. To that end, we constrain recipe embeddings so that intermediate representations of individual recipe components are close together when they belong to the same recipe, and far apart otherwise. For example, given the title representation of a recipe $e^{n=i}_{ttl}$ we define an objective function to make it closer to its corresponding ingredient representation $e^{n=i}_{ing}$ and farther from the representation of ingredients from other recipes $e^{n \neq i}_{ing}$. Formally, during training we incorporate a triplet loss term between title, ingredient and instruction embeddings that is defined as follows:
\begin{equation}
\mathcal{L}_{rec}'(a,b) = \mathcal{L}_{bi}(e_{a}^{n = i}, \hat{e}_{b \rightarrow a}^{n = i}, \hat{e}_{b \rightarrow  a}^{n\neq i}, e_{a}^{n \neq i})
\end{equation}
where $a$ and $b$ can both take values among the three different recipe components (title, ingredient and instructions). For every pair of values for $a$ and $b$, the embedding feature ${e}_{b}$ is projected to another feature space as $\hat{e}_{b \rightarrow a}$ using a single linear layer $g_{b \rightarrow a}(e_{b})$. Figure \ref{fig:recipe_loss} shows the 6 different projection functions for all possible combinations of $a$ and $b$. Note that, similarly to previous works in the context of self-supervised learning \cite{stroud2020learning} and learning from noisy data \cite{chen2020simple}, we optimise the loss between $e_{a}$ and $\hat{e}_{b \rightarrow a}$, instead of between $e_{a}$ and $e_b$. The motivation for this design is to leverage the shared semantics between recipe components, while still keeping the unique information that each component brings (i.e. information that is present in the ingredients might be missing in the title). By adding a projection before computing the loss, we enforce embeddings to be similar but not the same, avoiding the trivial solution of making all embeddings equal.%, which can hurt the retrieval performance.

We compute the loss above for all possible combinations of $a$ and $b$, and average the result: 

\begin{equation}
\mathcal{L}_{rec} =\dfrac{1}{6} \sum_{a}^{}\sum_{b}^{} \mathcal{L}_{rec}'(a,b)\delta(a,b), \label{eq:ssloss}
\end{equation}

where $a,b \in \{ttl, ing, ins\}$. Figure \ref{fig:recipe_loss} depicts the 6 different loss terms computed between all possible combinations of recipe components.

%Our experiments show that the addition of $L_{rec}$ boosts retrieval performance of our model, and allows us to expand the training set using recipe-only samples. Further, we demonstrate that $L_{rec}$ is also helpful when optimised using only paired data, showing that the self-supervised signal provided by relationships between recipe components is beneficial to obtain more powerful recipe representations, even without extra data.

%\textbf{Exploiting same-recipe food images}. Most works in the image-to-recipe literature exploit multiple images per recipe by sampling an image randomly during training. We argue that this approach may be suboptimal, since it does not enforce images that belong to the same recipe to be close to each other in the embedding space. We incorporate an additional triplet loss to account for that:
%
%\begin{equation}
%\mathcal{L}_{img} = \mathcal{L}(e_{I}^{n = i}, e_{I_2}^{n = i}, e_{I}^{n\neq i}, e_{I_2}^{n \neq i})
%\end{equation}

The final loss function is the composition of the paired loss and the recipe loss defined as: $\mathcal{L} = \alpha \mathcal{L}_{pair} +  \beta \mathcal{L}_{rec}$, where both $\alpha$ and $\beta$ are set to $1.0$ for paired samples, and $\alpha = 0.0$ and $\beta = 1.0$ for text-only samples.

%% file: sections/4_experiments.tex
\section{Experiments}
This section presents the experiments to validate the effectiveness of our proposed approach, including ablation studies, and comparison to previous works. 

%Sec.~\ref{ssec:details} presents the implementation details, the experimental setup and evaluation metrics. 
%We show the contribution of the proposed hierarchical transformer compared with other popular recipe encoder in Sec.~\ref{ssec:recipe_encoders}.
%Sec.~\ref{ssec:ablation} demonstrates that all recipe components are required to obtain state-of-the-art performance.
%Sec.~\ref{ssec:selfsupervised} provides evidence about our self-supervised recipe loss function.
%Sec.~\ref{ssec:incomplete_data} shows that our model works with incomplete recipe components by hallucinating the missing data. 
%We then compare our method to the state of the art in Sec.~\ref{ssec:sota} showing the superiority of our approach.
%To further improve our results, we experimented with different backbone networks for input images in Sec.~\ref{ssec:image_encoders}.
%Finally, we show qualitative results in Sec.~\ref{ssec:qualitative_results}.

\subsection{Implementation Details}
\label{ssec:details}
{\textbf{Dataset.} Following prior works, we use the Recipe1M \cite{salvador2017learning} dataset to train and evaluate our models. We use the official dataset splits which contain $238.999$, $51.119$ and $51.303$ image-recipe pairs for training, validation and testing, respectively. When we incorporate the self-supervised recipe loss from Section~\ref{ssec:additional_data}, we make use of the remaining part of the dataset that only contains recipes (no images), which adds $482.231$ samples that we use for training.

\textbf{Metrics.} Following previous works, we measure retrieval performance with the median rank (medR) and Recall@\{1, 5, 10\} (referred to as R1, R5, and R10). on rankings of size $ N = \{1.000, 10.000\}$. We report average metrics on $10$ groups of $N$ randomly chosen samples.% from the corresponding dataset split.

{\textbf{Training details.} We train models with a batch size of $128$ using the Adam~\cite{KingmaB14} optimiser with a base learning rate of $10^{-4}$ for all layers. We use step-wise learning rate decay of $0.1$ every 30 epochs and monitor validation $R@1$ every epoch, keeping the best model with respect to that metric for testing. Images are resized to $256$ pixels in their shortest side, and cropped to $224 \times 224$ pixels. During training, we take a random crop and horizontally flip the image with $0.5$ probability. At test time, we use center cropping. For experiments using text-only samples, we alternate mini-batches of paired and text-only data with a 1:1 ratio. In the case of recipe-only samples, we increase the batch size to $256$ to take advantage of the lower GPU memory requirements when dropping the image encoder.

\subsection{Recipe Encoders}
\label{ssec:recipe_encoders_exps}
We compare the proposed Transformer-based encoders from Section \ref{ssec:recipe_encoders} with their corresponding LSTM variants. We also quantify the gain of employing hierarchical versions by comparing them with simple average pooling on top of the outputs of a single sentence encoder (either LSTM or Transformer). Table \ref{tab:recipe_encoders} reports the results for the task of image-to-recipe retrieval in the validation set of Recipe1M. Transformers outperform LSTMs in both the averaging and hierarchical settings (referred to as \textit{+avg} and \textit{H-}) by 4.6 and 2.3 R1 points, respectively. Further, the use of hierarchical encoders provides a boost in performance with respect to the averaging baseline both for Transformers and LSTMs (increase of 4.2 and 1.9 R1 points, respectively). Given its favourable performance, we adopt the H-Transformer in the rest of the experiments. %TODO add parameters, comment on numbers

%Further, we also try a variant of the hierarchical Transformer in which the second layer consists of a single transformer that receives the embeddings of the title, ingredients and instructions (as opposed to a separate Transformer for ingredients and instructions).

\begin{table}[]
\centering
\begin{tabular}{@{}lrrrr@{}}
\toprule
                           & \multicolumn{1}{c}{\textbf{medR}} & \multicolumn{1}{c}{\textbf{R1}} & \multicolumn{1}{c}{\textbf{R5}} & \multicolumn{1}{c}{\textbf{R10}} \\ \midrule
LSTM + avg        & 9.0                                 & 17.9                                 & 41.2 & 52.9 \\
H-LSTM       & 7.0                               & 19.8                               & 44.8                             & 57.2                                  \\ 
Transformer + avg                     & 7.0                                 & 20.2                                  & 45.2                                 & 57.3                                \\ 
H-Transformer                             & \textbf{5.0}                                 & \textbf{24.4} & \textbf{51.4}                        & \textbf{63.4}  \\  \bottomrule
\end{tabular}
\caption{\textbf{Comparison between recipe encoders}. Image-to-recipe retrieval results reported on the validation set of Recipe1M. Results reported on rankings of size $10k$. }
\label{tab:recipe_encoders}
\vspace{-3mm}
\end{table}

\subsection{Ablation Study on Recipe Components}
\label{ssec:ablation}

In this section, we aim at quantifying the importance of each of the recipe components. Table~\ref{tab:ablation} reports image-to-recipe retrieval results for models trained and tested using different combinations of the recipe components.
Results in the first three rows indicate that the ingredients are the most important component, as the model achieves a R1 of $19.1$ when ingredients are used in isolation.
In contrast, R1 drops to $12.6$ and $6.0$ when using only the instructions and the title, respectively. Further, results improve when combining pairs of recipe components (rows 4-6), showing that using ingredients and instructions achieves the best performance of all possible pairs (R1 of $22.4$). Finally, the best performing model is the one using the full recipe (last row: R1 of $24.4$), suggesting that all recipe components contribute to the retrieval performance.
%We experimented with a single component (e.g., title only), pairs (e.g., title + ingredients) and the full recipe.

\begin{table}[]
\centering
\begin{tabular}{@{}lrrrr@{}}
\toprule
                           & \multicolumn{1}{c}{\textbf{medR}} & \multicolumn{1}{c}{\textbf{R1}} & \multicolumn{1}{c}{\textbf{R5}} & \multicolumn{1}{c}{\textbf{R10}} \\ \midrule
                           Ingredients only          & 8.2                              & 19.1                                 & 42.8                                 & 54.3                                  \\

Instructions only               & 15.0                                & 12.6                                 & 32.2                                 & 43.3                                  \\
Title  only                    & 35.5                               & 6.0                                 & 18.7                                 & 28.1                                  \\ \midrule
Ingrs + Instrs & 6.0                               & 22.4                               & 48.3 & 60.4 \\
Title + Ingrs        & 6.0                                 & 22.1                                 & 47.7 &59.8 \\
Title + Instrs       & 10.5                            & 15.9                              & 38.4 & 50.2 \\ \midrule
Full Recipe                     & \textbf{5.0}                                 & \textbf{24.4} & \textbf{51.4}                                 & \textbf{63.4}                                  \\
 \bottomrule
\end{tabular}
\caption{\textbf{Ablation studies of recipe components.} Image-to-recipe retrieval results reported on the validation set of Recipe1M. Results reported on rankings of size $10k$. }
\label{tab:ablation}
\vspace{-3mm}
\end{table}

\begin{table*}[]
\centering
\resizebox{\textwidth}{!}{
\begin{tabular}{@{}l|cccc|cccc|cccc|cccc@{}}
\toprule
\multirow{3}{*}{} & \multicolumn{8}{c|}{\textbf{1k}}                                                    & \multicolumn{8}{c}{\textbf{10k}}                                                   \\ \cmidrule(l){2-17} 
                  & \multicolumn{4}{c|}{\textbf{image-to-recipe}} & \multicolumn{4}{c|}{\textbf{recipe-to-image}} & \multicolumn{4}{c|}{\textbf{image-to-recipe}} & \multicolumn{4}{c}{\textbf{recipe-to-image}} \\ \cmidrule(l){2-17} 
                  & medR     & R1      & R5    & R10    & medR     & R1     & R5     & R10    & medR     & R1      & R5    & R10    & medR     & R1     & R5     & R10    \\ \midrule
Salvador et al. \cite{salvador2017learning} $^\diamond$   & 5.2      & 24.0    & 51.0      &   65.0     &   5.1       & 25.0       &     52.0   &       65.0 & 41.9    & -     &  -     &   -     &      39.2    &      -  &     -   &    -    \\
Chen et al. \cite{chen2018deep}       &    4.6      &    25.6     &   53.7    &    66.9    &    4.6      &    25.7       &   53.9     &         67.1 &  39.8       &    7.2   &     19.2   &     27.6     &  38.1   &  7.0 &   19.4     &   27.8     \\
Carvalho et al. \cite{carvalho2018cross} $^\diamond$ & 2.0      & 39.8    &  69.0     &    77.4    &    1.0      &   40.2     &   68.1     &   78.7     & 13.2     & 14.9    &   35.3    &  45.2      &      12.2    &   14.8     &    34.6    &    46.1    \\
R2GAN \cite{zhu2019r2gan}$^\diamond$                  &     2.0     &     39.1    &   71.0    &  81.7      &  2.0        &    40.6    &    72.6    &       83.3 &  13.9        &   13.5      &   33.5    &     44.9   &     12.6     &   14.2     &   35.0     &   46.8     \\
MCEN \cite{fu2020mcen}               &   2.0       &   48.2      &    75.8   &      83.6  &    1.9      &   48.4     &  76.1      &    83.7    &  7.2        &      20.3   &    43.3   &   54.4     &     6.6     &   21.4     &  44.3      &    55.2    \\
   ACME \cite{wang2019learning}$^\diamond$               &    1.0      &   51.8      &   80.2    &    87.5    &   1.0       &    52.8    &   80.2     &       87.6 &     6.7     &   22.9     &  46.8     &    57.9    &      6.0    &  24.4      &    47.9    &      59.0  \\
   SCAN \cite{wang2020cross}               &    1.0      & 54.0        &     81.7  &    88.8    &    1.0      &  54.9      &  81.9      &   89.0     &    5.9      &    23.7    &    49.3   &    60.6    &    5.1      &     25.3   &  50.6      &  61.6      \\
  
   DaC \cite{fain2019dividing}             &  -       &     -    &    -   &    -    &     -     &     -   &    -    &      -  &     5.9    &    24.4     &    49.4   &    60.5    &    -      &     -   &     -   &        \\
   DaC \cite{fain2019dividing}$^\diamond$              &  1.0        &     55.9    &    82.4   &    88.7    &   -       &     -   &    -    &        &     5.0    &    26.5     &    51.8   &    62.6    &    -      &    -    &    -    &  -      \\ \midrule
Ours ($\mathcal{L}_{pair}$)            &   1.0      &    58.3    &    86.2   &  91.8      &  1.0       &   59.6  &   86.1     &     92.2   &   4.1      &    26.8     &   54.7   &    66.5    &   4.0       &   27.6     &     55.1   &   66.8    \\
Ours ($\mathcal{L}_{pair}+\mathcal{L}_{rec}$)        &    1.0      &    59.1     &    86.9   &   92.3     &   1.0       &   59.1     &    87.0   &    92.7  &     4.0                                & 27.3                                 & 55.4 & 67.3 &    4.0      &    27.8    &  55.6     &   67.3  \\
Ours $(\mathcal{L}_{pair}+\mathcal{L}_{rec})^\diamond$             &       1.0   &    \textbf{60.0}     &    \textbf{87.6}   &    \textbf{92.9}    &     1.0     &    \textbf{60.3}    &   \textbf{87.6}    &   \textbf{93.2}     &     4.0     &    \textbf{27.9}     &   \textbf{56.4}    &        \textbf{68.1}&   4.0    &  \textbf{28.3}      &    \textbf{56.5}    &   \textbf{68.1}     \\  \bottomrule
\end{tabular}
}
\caption{\textbf{Comparison with existing methods.} medR ($\downarrow$), Recall@k ($\uparrow$) are reported on the Recipe1M test set. $\diamond$ indicates that methods use all training samples in Recipe1M for training as opposed to using paired image-recipe samples only. }
\label{tab:sotacomparison}
\end{table*}

\subsection{Self-supervised Recipe Loss}
\label{ssec:selfsupervised}

With the goal of understanding the contribution of the self-supervised loss described in Section~\ref{ssec:additional_data}, we compare the performance of three model variants in the last three rows of Table \ref{tab:sotacomparison}:
$\mathcal{L}_{pair}$ only uses the loss function for paired image-recipe data,
$\mathcal{L}_{pair}+\mathcal{L}_{rec}$ adds the self-supervised loss considering only paired data, and
$(\mathcal{L}_{pair}+\mathcal{L}_{rec})^\diamond$ is trained on both paired and recipe-only samples. The self-supervised learning approach $\mathcal{L}_{pair}+\mathcal{L}_{rec}$ improves performance with respect to $\mathcal{L}_{pair}$, while using the same amount of paired data (improvement of 0.5 R1 points on the image-to-recipe setting for rankings of size 10k). These results indicate that enforcing a similarity between pairs of recipe components helps to make representations more robust, leading to better performance even without extra training data. The last row of Table \ref{tab:sotacomparison} shows the performance of $(\mathcal{L}_{pair}+\mathcal{L}_{rec})^\diamond$, which is trained with the addition of the self-supervised loss, optimised for both paired and recipe-only data. Significant improvements for image-to-recipe retrieval are obtained for both median rank and recall metrics with respect to $\mathcal{L}_{pair}$: medR decreases to $4.1$ from $4.0$ and R1 lifts up from $26.8$ to $27.9$. These results indicate that both the self-supervised loss term and the additional training data contribute to the performance improvement. We also quantify the contribution of the $g(\cdot)$ functions from Figure \ref{fig:recipe_loss} by comparing to a baseline model in which they are replaced with identity functions. This model achieves slightly worse retrieval results with respect to $(\mathcal{L}_{pair}+\mathcal{L}_{rec})^\diamond$ (0.5 point decrease in terms of R1).

\subsection{Comparison to existing works}
\label{ssec:sota}

%When available, we report the performance of existing methods using their publicly available code and pretrained models. Otherwise, we provide numbers reported by authors.

We compare the performance of our method with existing works in Table \ref{tab:sotacomparison}, where we take our best performing model on the validation set, and evaluate its performance on the test set. For comparison, we provide numbers reported by authors in their respective papers. When trained with paired data only, our model $\mathcal{L}_{pair} + \mathcal{L}_{rec}$ achieves the best results compared to recent methods trained using the same data, achieving an image-to-recipe R1 of 27.3 on 10k-sized rankings (c.f. 24.4 DaC \cite{fain2019dividing}, 23.7 SCAN \cite{wang2020cross}, and 20.3 MCEN \cite{fu2020mcen}). When we incorporate the additional unpaired data with no images, it makes a further improvement in the retrieval accuracy (R1 of 27.9, and R5 of 56.4), while still outperforming the state-of-the-art method of DaC$^\diamond$ \cite{fain2019dividing}, which jointly embeds pre-trained recipe embeddings (trained on the full training set) and pre-trained image representations using triplet loss. Compared to previous works, we use raw recipe data as input (as opposed to using partial recipe information, or pre-trained embeddings), and train the model with a simple loss functions that are directly applied to the output embeddings and intermediate representations. Our model ($\mathcal{L}_{pair} + \mathcal{L}_{rec})^\diamond$ achieves state-of-the-art results for all retrieval metrics (medR and recall) and retrieval scenarios (image-to-recipe and recipe-to-image) for 10k-sized rankings, while being conceptually simpler and easier to train both in terms of data preparation and optimization compared previous works.%Our method's variants achieve favourable performance against recent state of the art methods in terms of all settings, i.e. medR and $R\{1, 5, 10\}$ in the $1k$ and $10k$ retrieval.

%Our model outperforms \cite{zhu2019r2gan}, which uses surrogate adversarial losses for image and recipe reconstruction from joint embeddings, as well as semantic regularization losses. In both retrieval settings, our model reaches similar performance to \cite{wang2019learning}, which uses pre-trained embeddings for recipe components, as well as adversarial losses for translation consistency. 

\subsection{Testing with incomplete data}
\label{ssec:incomplete_data}

\begin{table}[]
\centering
\begin{tabular}{@{}lrrrr@{}}
\toprule
                           & \multicolumn{1}{c}{\textbf{medR}} & \multicolumn{1}{c}{\textbf{R1}} & \multicolumn{1}{c}{\textbf{R5}} & \multicolumn{1}{c}{\textbf{R10}} \\ \midrule
No title                    & 6.0                                & 22.7                                  &48.4 & 60.4                                  \\ 
Hallucinated $e_{ttl}$                              &5.0                                 & 24.2                               & 51.2      & 63.1                            \\ \midrule
No ingredients       & 10.2                                 & 16.0                                   & 38.3 & 50.2 \\
Hallucinated $e_{ing}$       & 10.1                         & 16.6                               & 39.1                                 & 50.8                                  \\ \midrule
No instructions         & 6.0                             & 22.3                            & 48.0 & 59.8                                  \\ 
Hallucinated  $e_{ins}$          & 6.0                              & 23.1                                & 49.4 & 61.1                                  \\ \midrule
Title only                    & 35.5                                & 6.0                                 &18.9 & 28.4                                 \\ 
Hallucinated $e_{ing}$, $e_{ins}$                                &35.8                                & 6.6                              &20.0     & 29.3 \\ \midrule
Ingredients only      & 8.3                                 & 19.2                                & 42.5 & 53.9 \\
Hallucinated $e_{ttl}$, $e_{ins}$         & 8.0                        & 19.4                               & 43.5                                 & 55.3                                  \\ \midrule
Instructions only         & 15.0                            & 13.1                           & 32.6 & 43.8                                 \\ 
Hallucinated  $e_{ttl}$, $e_{ing}$          & 13.9                              & 14.0                               & 34.1 & 45.4 \\ \bottomrule
\end{tabular}
\caption{\textbf{Testing with missing data.} Image-to-recipe retrieval results reported on the test set of Recipe1M. Results reported on rankings of size $10k$. }
\label{tab:missing_data}
\end{table}

\begin{table}[]
\centering
\begin{tabular}{@{}lrrrr@{}}
\toprule
                           & \multicolumn{1}{c}{\textbf{medR}} & \multicolumn{1}{c}{\textbf{R1}} & \multicolumn{1}{c}{\textbf{R5}} & \multicolumn{1}{c}{\textbf{R10}} \\ \midrule
DaC (ResNeXt-101)  \cite{fain2019dividing}      & 4.0                               & 30.0                                & 56.5                                 & 67.0                                  \\ \midrule
%ResNet-50                     & 4.1     &    26.7     &   54.3    &        65.8                               \\  
ResNet-50                              & 4.0                                & 27.9                                 & 56.4 &68.1 \\
ResNeXt-101       & 4.0                                & 28.9                                & 57.4 & 69.0 \\
ViT      & \textbf{3.0}                               & \textbf{33.5}                                 & \textbf{62.2}                            & \textbf{72.9} \\ 
 \bottomrule
\end{tabular}
\caption{\textbf{Comparison of different image encoders.} Image-to-recipe retrieval results reported on the test set of Recipe1M. Results reported on rankings of size $10k$. }
\label{tab:image_encoders}
\vspace{-5mm}
\end{table}

 \begin{figure}[h]
    \centering

      \includegraphics[trim=1100 150 800 150,clip, width=\columnwidth]{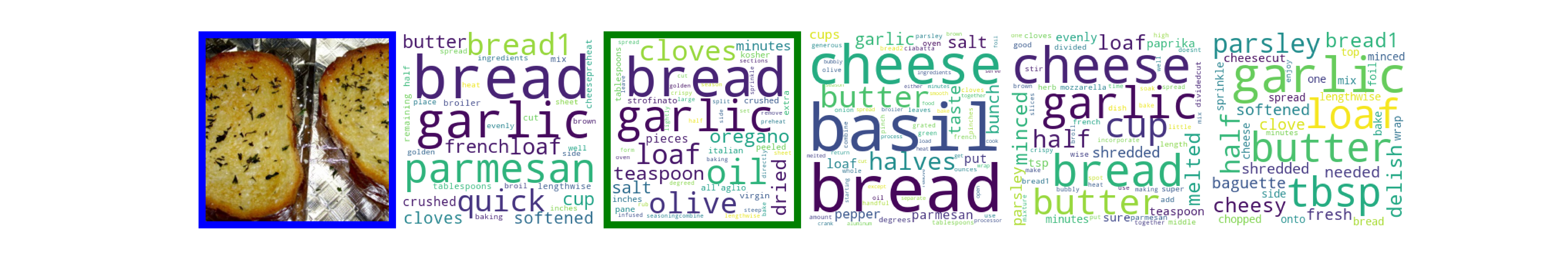}
       \includegraphics[trim=1100 150 800 150, clip, width=\columnwidth]{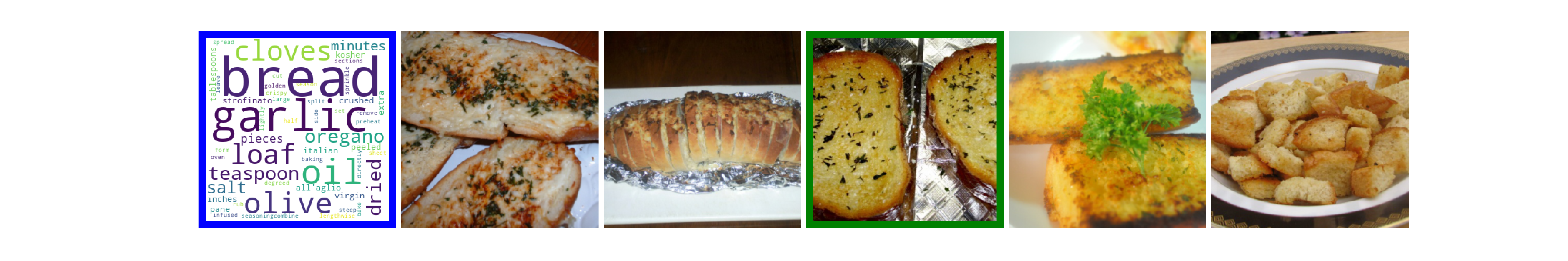}
        \includegraphics[trim=1100 150 800 150,clip, width=\columnwidth]{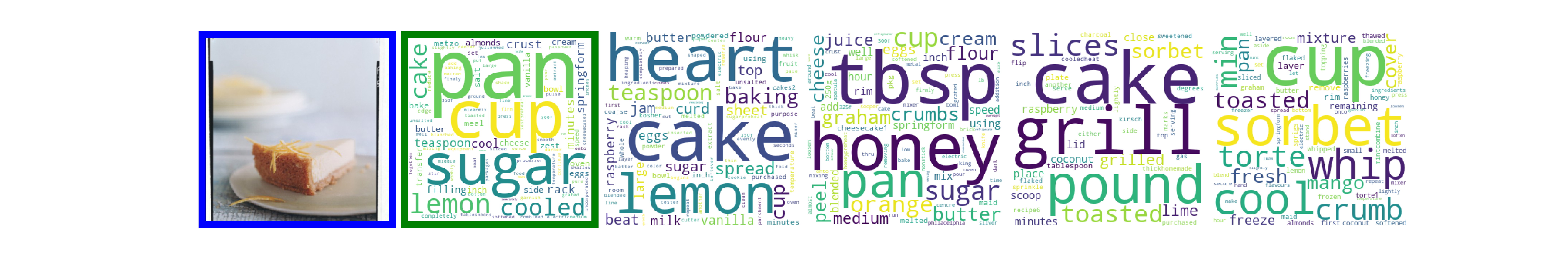}
         \includegraphics[trim=1100 150 800 150,clip, width=\columnwidth]{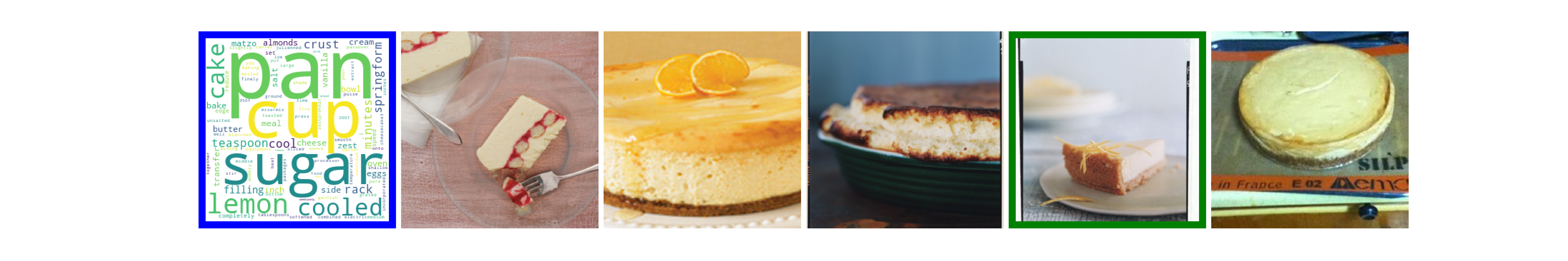}
          \includegraphics[trim=1100 150 800 150,clip, width=\columnwidth]{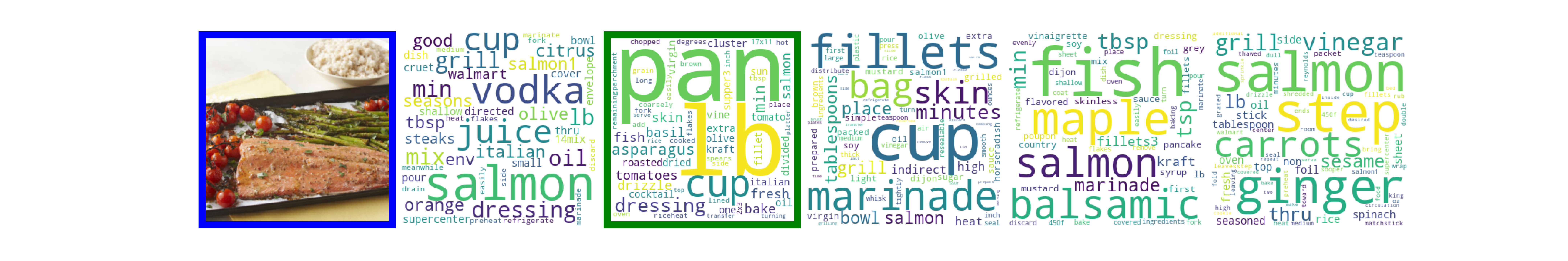}
           \includegraphics[trim=1100 150 800 150,clip, width=\columnwidth]{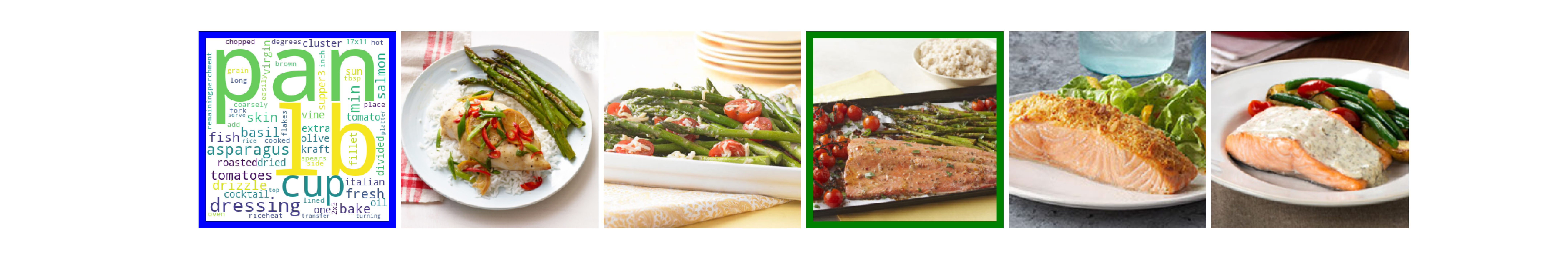}
    \caption{\textbf{Qualitative results.} Each row includes the query (image or recipe) on the left (highlighted in blue), followed by the top $K=5$ retrieved recipes. The correct retrieved element is highlighted in green.}

    \label{fig:qualitativeresults}
    \vspace{-4mm}
\end{figure}

Training with our self-supervised triplet loss on recipe components allows us to easily test our model in missing data scenarios.
Once trained, our proposed projection layers described in Section~\ref{ssec:additional_data} allow our model to hallucinate any recipe component from the others, e.g. in case that the title is missing, we can simply take the average of the two respective projected vectors from the ingredients and the instructions: $e_{ttl}$ as $(g_{ing\rightarrow ttl}(e_{ing}) + g_{ins\rightarrow ttl}(e_{ins}))/2$ (see Figure~\ref{fig:recipe_loss} for reference). We pick the model trained with $\mathcal{L}_{pair} + \mathcal{L}_{rec}$ and evaluate its image-to-recipe performance when some recipe component features are replaced with their hallucinated versions. In Table \ref{tab:missing_data}, we compare models using hallucinated features with respect to the ones in Table \ref{tab:ablation}, i.e. those that ignore those inputs completely during training. In all missing data combinations, we see a consistent improvement over the cases where the missing data is not used during training. Results suggest that using all recipe components during training can improve performance even when some of them are missing at test time.% Moreover, our model can still achieve comparable results to the state of the art with missing data (\emph{c.f.} \textit{~No title} case in Table \ref{tab:missing_data}). 

%At this point, one can argue that using a shared textual encoder for each recipe component and/or averaging embeddings can also provide robustness to missing data. However, we experimentally observed inferior results in the complete data and incomplete data cases for those simpler approaches.

%\begin{figure}[h]
%    \centering
%    \includegraphics[width=\columnwidth]{figures/rank_histogram.png}
%    \caption{Histogram of rank distribution for different methods.}
%    \label{fig:rank_histogram}
%\end{figure}

\begin{figure}[h]
    \centering
%  \includegraphics[trim=900 150 700 0,clip, width=\columnwidth]{figures/comparison/0ca9bd3da2.png}
%    % \includegraphics[trim=900 150 700 0,clip, width=\columnwidth]{figures/comparison/0f6e15996d.png}
%      \includegraphics[trim=900 150 700 0,clip, width=\columnwidth]{figures/comparison/6cda1f3508.png}
%      % \includegraphics[trim=900 150 700 0, clip, width=\columnwidth]{figures/comparison/6ec19e6124.png}
%        \includegraphics[trim=900 150 700 0,clip, width=\columnwidth]{figures/comparison/97d846093a.png}
%        \includegraphics[trim=900 150 700 0,clip, width=\columnwidth]{figures/comparison/615e73ba74.png}
	\includegraphics[width=0.9\columnwidth]{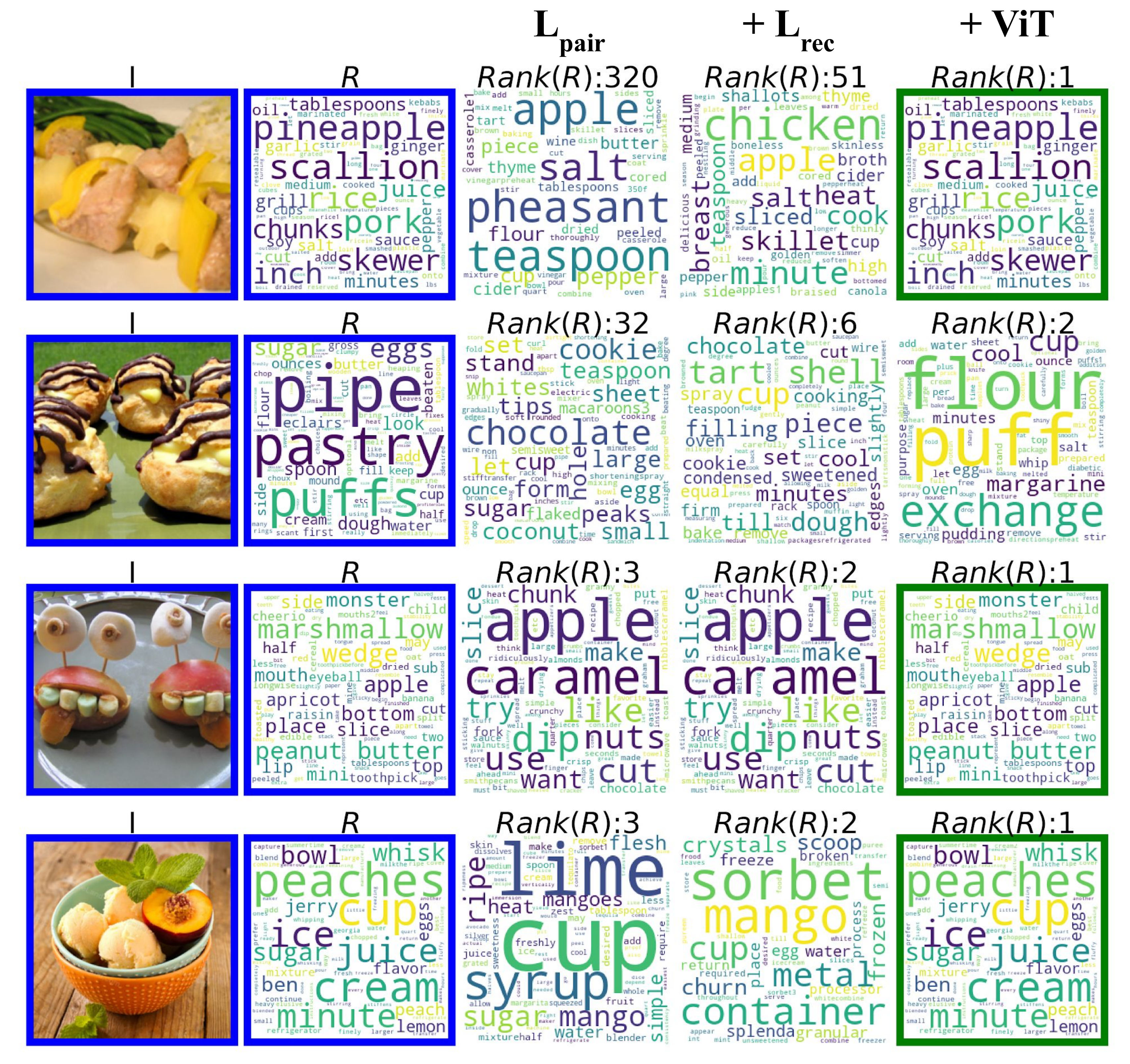}

    \caption{\textbf{Incremental improvements.} Each row includes top-1 retrieved recipes for different methods. From left to right: a) Query Image, b) True Recipe, c) $\mathcal{L}_{pair}$, d) $(\mathcal{L}_{pair}$+ $\mathcal{L}_{rec}) ^\diamond$, and e) ($\mathcal{L}_{pair}$+ $\mathcal{L}_{rec}) ^\diamond$ (ViT).}

    \label{fig:comparison}
    \vspace{-3mm}
\end{figure}
\subsection{Image Encoders}
\label{ssec:image_encoders}

We report the performance of our best model ($\mathcal{L}_{pair} + \mathcal{L}_{rec}$)$^\diamond$ using different image encoders in Table \ref{tab:image_encoders}. For comparison with \cite{fain2019dividing}, we train our model with ResNeXt-101 image encoder. For $R\{5, 10\}$, we achieve favourable performance with respect to \cite{fain2019dividing} while sharing the same medR score when using the same encoder. We also experiment with the recently introduced Visual Transformer (ViT) \cite{dosovitskiy2020image} as image encoder, and achieved substantial improvement for all metrics: medR of 3.0 and $R\{1, 5, 10\}$ improvement of 3.5, 5.7 and 5.9 points, respectively compared to the best reported results so far on Recipe1M (Table \ref{tab:image_encoders}, row 1).

\subsection{Qualitative results} 
\label{ssec:qualitative_results}

%\AS{TODO: more interesting visualizations. eg off-by-one examples, fine grained examples, failure cases}
Figure \ref{fig:qualitativeresults} shows some qualitative image-to-recipe and recipe-to-image retrieval using our learned embeddings using the best performing model from Table \ref{tab:image_encoders}\footnote{Recipes are shown as world clouds (\url{https://amueller.github.io/word_cloud/}) for simplicity.}}. Our model is able to find recipes that include relevant ingredient words to the query food image (\emph{e.g.} \emph{bread} and \emph{garlic} in the first row, \emph{salmon} in the fifth row). Figure \ref{fig:comparison} shows examples of the performance improvement of our different models. When adding our proposed recipe loss $\mathcal{L}_{rec}$, and replacing the image model with ViT, the rank of the correct recipe ($Rank(R)$) is improved, as well as the relevance of the top retrieved recipe with respect to the correct one in terms of the common words. These results indicate that our proposed model not only improves retrieval accuracy, but also returns more semantically similar recipes with respect to the query.

%Figure \ref{fig:qualitativeresults} shows some qualitative results of our learned embeddings on the image-to-recipe retrieval tasks. While we perform cross-modal retrieval in the joint embedding space (i.e. recipes are ranked based on the cosine similarities between the query image vector $e_I$ and all database recipe vectors $e_R^{n=1:N}$), for the purposes of visualization we display the image associated to each of the retrieved recipes in the ranking.
 %trim=0 0 0 0,clip, 

%% file: sections/5_conclusions.tex
\section{Conclusion}

In this work, we study the cross-modal retrieval problem in the food domain by addressing different limitations from previous works. We first propose a textual representation model based on hierarchical Transformers outperforming LSTM-based recipe encoders. Secondly, we propose a self-supervised loss to account for relations between different recipe components, which is straightforward to add on top of intermediate recipe representations, and significantly improves the retrieval results. Moreover, this loss allows us to train using both paired and unpaired recipe data (i.e. recipes without images), resulting in further boost in performance. As a result of our contributions, our method achieves state-of-the-art results in the Recipe1M dataset.

%Additional advantage of our projection layers consumed in the self-supervised loss is that it enables to hallucinate the missing recipe components in their absence, maintaining comparable performance to the state-of-the-art. 
%We presented a simple architecture to learn joint embeddings from food images and textual recipes, that achieves competitive performance on the recipe retrieval task compared to existing works. In contrast to previous methods, which use pre-trained text representations and auxiliary loss functions to train their joint recipe-image embedding, our model is trained end-to-end on raw recipe data using a single triplet-loss based objective function.

%% file: cvpr.bbl
\begin{thebibliography}{10}\itemsep=-1pt

\bibitem{amac2019procedural}
Mustafa~Sercan Amac, Semih Yagcioglu, Aykut Erdem, and Erkut Erdem.
\newblock Procedural reasoning networks for understanding multimodal
  procedures.
\newblock {\em arXiv preprint arXiv:1909.08859}, 2019.

\bibitem{bossard2014food}
Lukas Bossard, Matthieu Guillaumin, and Luc Van~Gool.
\newblock Food-101--mining discriminative components with random forests.
\newblock In {\em ECCV}, 2014.

\bibitem{carvalho2018cross}
Micael Carvalho, R{\'e}mi Cad{\`e}ne, David Picard, Laure Soulier, Nicolas
  Thome, and Matthieu Cord.
\newblock Cross-modal retrieval in the cooking context: Learning semantic
  text-image embeddings.
\newblock In {\em ACM SIGIR Conference on Research \& Development in
  Information Retrieval}, 2018.

\bibitem{chandu2019storyboarding}
Khyathi Chandu, Eric Nyberg, and Alan~W Black.
\newblock Storyboarding of recipes: Grounded contextual generation.
\newblock In {\em ACL}, 2019.

\bibitem{chen2016deep}
Jing-Jing Chen and Chong-Wah Ngo.
\newblock Deep-based ingredient recognition for cooking recipe retrieval.
\newblock In {\em ACM MM}. ACM, 2016.

\bibitem{chen2017cross}
Jing-Jing Chen, Chong-Wah Ngo, and Tat-Seng Chua.
\newblock Cross-modal recipe retrieval with rich food attributes.
\newblock In {\em ACM MM}. ACM, 2017.

\bibitem{chen2018deep}
Jing-Jing Chen, Chong-Wah Ngo, Fu-Li Feng, and Tat-Seng Chua.
\newblock Deep understanding of cooking procedure for cross-modal recipe
  retrieval.
\newblock In {\em ACM MM}, 2018.

\bibitem{Chen2012quantities}
Mei-Yun Chen, Yung-Hsiang Yang, Chia-Ju Ho, Shih-Han Wang, Shane-Ming Liu,
  Eugene Chang, Che-Hua Yeh, and Ming Ouhyoung.
\newblock Automatic chinese food identification and quantity estimation.
\newblock In {\em SIGGRAPH Asia 2012 Technical Briefs}, 2012.

\bibitem{chen2020simple}
Ting Chen, Simon Kornblith, Mohammad Norouzi, and Geoffrey Hinton.
\newblock A simple framework for contrastive learning of visual
  representations.
\newblock {\em arXiv preprint arXiv:2002.05709}, 2020.

\bibitem{chen2015microsoft}
Xinlei Chen, Hao Fang, Tsung-Yi Lin, Ramakrishna Vedantam, Saurabh Gupta, Piotr
  Doll{\'a}r, and C~Lawrence Zitnick.
\newblock Microsoft coco captions: Data collection and evaluation server.
\newblock {\em arXiv preprint arXiv:1504.00325}, 2015.

\bibitem{chen2017chinesefoodnet}
Xin Chen, Hua Zhou, and Liang Diao.
\newblock Chinesefoodnet: {A} large-scale image dataset for chinese food
  recognition.
\newblock {\em CoRR}, abs/1705.02743, 2017.

\bibitem{dosovitskiy2020image}
Alexey Dosovitskiy, Lucas Beyer, Alexander Kolesnikov, Dirk Weissenborn,
  Xiaohua Zhai, Thomas Unterthiner, Mostafa Dehghani, Matthias Minderer, Georg
  Heigold, Sylvain Gelly, et~al.
\newblock An image is worth 16x16 words: Transformers for image recognition at
  scale.
\newblock {\em arXiv preprint arXiv:2010.11929}, 2020.

\bibitem{fain2019dividing}
Mikhail Fain, Andrey Ponikar, Ryan Fox, and Danushka Bollegala.
\newblock Dividing and conquering cross-modal recipe retrieval: from nearest
  neighbours baselines to sota.
\newblock {\em arXiv preprint arXiv:1911.12763}, 2019.

\bibitem{fu2020mcen}
Han Fu, Rui Wu, Chenghao Liu, and Jianling Sun.
\newblock Mcen: Bridging cross-modal gap between cooking recipes and dish
  images with latent variable model.
\newblock In {\em CVPR}, 2020.

\bibitem{gu2018look}
Jiuxiang Gu, Jianfei Cai, Shafiq~R Joty, Li Niu, and Gang Wang.
\newblock Look, imagine and match: Improving textual-visual cross-modal
  retrieval with generative models.
\newblock In {\em CVPR}, 2018.

\bibitem{he2016deep}
Kaiming He, Xiangyu Zhang, Shaoqing Ren, and Jian Sun.
\newblock Deep residual learning for image recognition.
\newblock In {\em CVPR}, 2016.

\bibitem{lstm}
Sepp Hochreiter and J{\"u}rgen Schmidhuber.
\newblock Long short-term memory.
\newblock {\em Neural computation}, 1997.

\bibitem{huang2019acmm}
Yan Huang and Liang Wang.
\newblock Acmm: Aligned cross-modal memory for few-shot image and sentence
  matching.
\newblock In {\em ICCV}, 2019.

\bibitem{Karpathy:2017}
Andrej Karpathy and Li Fei-Fei.
\newblock Deep visual-semantic alignments for generating image descriptions.
\newblock In {\em CVPR}, 2015.

\bibitem{KingmaB14}
Diederik~P. Kingma and Jimmy Ba.
\newblock Adam: {A} method for stochastic optimization.
\newblock {\em CoRR}, abs/1412.6980, 2014.

\bibitem{kiros2015skip}
Ryan Kiros, Yukun Zhu, Russ~R Salakhutdinov, Richard Zemel, Raquel Urtasun,
  Antonio Torralba, and Sanja Fidler.
\newblock Skip-thought vectors.
\newblock In {\em NeurIPS}, 2015.

\bibitem{krizhevsky2012imagenet}
Alex Krizhevsky, Ilya Sutskever, and Geoffrey~E Hinton.
\newblock Imagenet classification with deep convolutional neural networks.
\newblock In {\em NeurIPS}, 2012.

\bibitem{Lee_2018_CVPR}
Kuang-Huei Lee, Xiaodong He, Lei Zhang, and Linjun Yang.
\newblock Cleannet: Transfer learning for scalable image classifier training
  with label noise.
\newblock In {\em CVPR}, 2018.

\bibitem{li2020picture}
Jiatong Li, Fangda Han, Ricardo Guerrero, and Vladimir Pavlovic.
\newblock Picture-to-amount (pita): Predicting relative ingredient amounts from
  food images.
\newblock {\em arXiv preprint arXiv:2010.08727}, 2020.

\bibitem{Liu2016DDL}
Chang Liu, Yu Cao, Yan Luo, Guanling Chen, Vinod Vokkarane, and Yunsheng Ma.
\newblock Deepfood: Deep learning-based food image recognition for
  computer-aided dietary assessment.
\newblock In {\em ICOST}, 2016.

\bibitem{liu2019hierarchical}
Yang Liu and Mirella Lapata.
\newblock Hierarchical transformers for multi-document summarization.
\newblock {\em ACL}, 2019.

\bibitem{im2calories}
Austin Meyers, Nick Johnston, Vivek Rathod, Anoop Korattikara, Alex Gorban,
  Nathan Silberman, Sergio Guadarrama, George Papandreou, Jonathan Huang, and
  Kevin~P Murphy.
\newblock Im2calories: towards an automated mobile vision food diary.
\newblock In {\em ICCV}, 2015.

\bibitem{nu9070657}
Simon Mezgec and Barbara Koroušić~Seljak.
\newblock Nutrinet: A deep learning food and drink image recognition system for
  dietary assessment.
\newblock {\em Nutrients}, 9(7), 2017.

\bibitem{mikolov2013efficient}
Tomas Mikolov, Kai Chen, Greg Corrado, and Jeffrey Dean.
\newblock Efficient estimation of word representations in vector space.
\newblock {\em ICLR}, 2013.

\bibitem{min2019survey}
Weiqing Min, Shuqiang Jiang, Linhu Liu, Yong Rui, and Ramesh Jain.
\newblock A survey on food computing.
\newblock {\em ACM Computing Surveys (CSUR)}, 52(5):1--36, 2019.

\bibitem{min2020isia}
Weiqing Min, Linhu Liu, Zhiling Wang, Zhengdong Luo, Xiaoming Wei, Xiaolin Wei,
  and Shuqiang Jiang.
\newblock Isia food-500: A dataset for large-scale food recognition via stacked
  global-local attention network.
\newblock In {\em ACM MM}, 2020.

\bibitem{Ngo2017DLF}
Chong-Wah Ngo.
\newblock Deep learning for food recognition.
\newblock In {\em SoICT}, 2017.

\bibitem{nishimura2019procedural}
Taichi Nishimura, Atsushi Hashimoto, and Shinsuke Mori.
\newblock Procedural text generation from a photo sequence.
\newblock In {\em Proceedings of the 12th International Conference on Natural
  Language Generation}, 2019.

\bibitem{ofli2017saki}
Ferda Ofli, Yusuf Aytar, Ingmar Weber, Raggi al Hammouri, and Antonio Torralba.
\newblock Is saki\# delicious?: The food perception gap on instagram and its
  relation to health.
\newblock In {\em ICWWW}, 2017.

\bibitem{pan2020chefgan}
Siyuan Pan, Ling Dai, Xuhong Hou, Huating Li, and Bin Sheng.
\newblock Chefgan: Food image generation from recipes.
\newblock In {\em ACM MM}, 2020.

\bibitem{salvador2019inverse}
Amaia Salvador, Michal Drozdzal, Xavier Giro-i Nieto, and Adriana Romero.
\newblock Inverse cooking: Recipe generation from food images.
\newblock In {\em CVPR}, 2019.

\bibitem{salvador2017learning}
Amaia Salvador, Nicholas Hynes, Yusuf Aytar, Javier Marin, Ferda Ofli, Ingmar
  Weber, and Antonio Torralba.
\newblock Learning cross-modal embeddings for cooking recipes and food images.
\newblock In {\em CVPR}, 2017.

\bibitem{sharma2018conceptual}
Piyush Sharma, Nan Ding, Sebastian Goodman, and Radu Soricut.
\newblock Conceptual captions: A cleaned, hypernymed, image alt-text dataset
  for automatic image captioning.
\newblock In {\em ACL}, 2018.

\bibitem{stroud2020learning}
Jonathan~C Stroud, David~A Ross, Chen Sun, Jia Deng, Rahul Sukthankar, and
  Cordelia Schmid.
\newblock Learning video representations from textual web supervision.
\newblock {\em arXiv preprint arXiv:2007.14937}, 2020.

\bibitem{vaswani2017attention}
Ashish Vaswani, Noam Shazeer, Niki Parmar, Jakob Uszkoreit, Llion Jones,
  Aidan~N Gomez, {\L}ukasz Kaiser, and Illia Polosukhin.
\newblock Attention is all you need.
\newblock In {\em NeurIPS}, 2017.

\bibitem{wang2020structure}
Hao Wang, Guosheng Lin, Steven~CH Hoi, and Chunyan Miao.
\newblock Structure-aware generation network for recipe generation from images.
\newblock {\em ECCV}, 2020.

\bibitem{wang2019learning}
Hao Wang, Doyen Sahoo, Chenghao Liu, Ee-peng Lim, and Steven~CH Hoi.
\newblock Learning cross-modal embeddings with adversarial networks for cooking
  recipes and food images.
\newblock In {\em CVPR}, 2019.

\bibitem{wang2020cross}
Hao Wang, Doyen Sahoo, Chenghao Liu, Ke Shu, Palakorn Achananuparp, Ee-peng
  Lim, and Steven~CH Hoi.
\newblock Cross-modal food retrieval: Learning a joint embedding of food images
  and recipes with semantic consistency and attention mechanism.
\newblock {\em arXiv preprint arXiv:2003.03955}, 2020.

\bibitem{wang2015recipe}
Xin Wang, Devinder Kumar, Nicolas Thome, Matthieu Cord, and Frederic Precioso.
\newblock Recipe recognition with large multimodal food dataset.
\newblock In {\em ICMEW}, 2015.

\bibitem{xie2017aggregated}
Saining Xie, Ross Girshick, Piotr Doll{\'a}r, Zhuowen Tu, and Kaiming He.
\newblock Aggregated residual transformations for deep neural networks.
\newblock In {\em CVPR}, 2017.

\bibitem{yagcioglu2018recipeqa}
Semih Yagcioglu, Aykut Erdem, Erkut Erdem, and Nazli Ikizler-Cinbis.
\newblock Recipeqa: A challenge dataset for multimodal comprehension of cooking
  recipes.
\newblock {\em arXiv preprint arXiv:1809.00812}, 2018.

\bibitem{young2014image}
Peter Young, Alice Lai, Micah Hodosh, and Julia Hockenmaier.
\newblock From image descriptions to visual denotations: New similarity metrics
  for semantic inference over event descriptions.
\newblock {\em Transactions of the Association for Computational Linguistics},
  2, 2014.

\bibitem{zhang-etal-2019-hibert}
Xingxing Zhang, Furu Wei, and Ming Zhou.
\newblock Hibert: Document level pre-training of hierarchical bidirectional
  transformers for document summarization.
\newblock {\em ACL}, 2019.

\bibitem{zhu2020cookgan}
Bin Zhu and Chong-Wah Ngo.
\newblock Cookgan: Causality based text-to-image synthesis.
\newblock In {\em CVPR}, 2020.

\bibitem{zhu2019r2gan}
Bin Zhu, Chong-Wah Ngo, Jingjing Chen, and Yanbin Hao.
\newblock R2gan: Cross-modal recipe retrieval with generative adversarial
  network.
\newblock In {\em CVPR}, 2019.

\end{thebibliography}
